\documentclass{article}

    \PassOptionsToPackage{numbers, compress}{natbib}
\usepackage[preprint]{neurips_data_2024}
\usepackage[utf8]{inputenc} % allow utf-8 input
\usepackage[T1]{fontenc}    % use 8-bit T1 fonts
\usepackage{hyperref}       % hyperlinks
\usepackage{url}            % simple URL typesetting
\usepackage{booktabs}       % professional-quality tables
\usepackage{amsfonts}       % blackboard math symbols
\usepackage{nicefrac}       % compact symbols for 1/2, etc.
\usepackage{microtype}      % microtypography
\usepackage{xcolor}         % colors

\usepackage{graphicx}
\usepackage{amsmath,amssymb,amsthm,amsfonts}
\usepackage{bm,bbm}
\usepackage{acronym}
\usepackage{enumitem}
\usepackage{balance}
\usepackage{xspace}
\usepackage[skip=3pt]{subcaption}
\usepackage[skip=3pt]{caption}
\usepackage{url}
\usepackage{multirow}
\usepackage{booktabs}
\usepackage{marvosym}
\usepackage{mathrsfs}
\usepackage{dashrule}
\usepackage{arydshln}
\usepackage{array}
\usepackage{colortbl}
\usepackage{pifont}
\newcommand\cmark {\textcolor{green}{\ding{52}}}
\newcommand\xmark {\textcolor{red}{\ding{55}}}

\newcommand{\model}{TLV-Link\xspace}
\newcommand{\dataset}{Touch100k\xspace}

\usepackage{makecell}

\definecolor{myblue}{RGB}{0,105,252}
\definecolor{myred}{RGB}{247,96,102}
\definecolor{mygray}{gray}{0.4}
\definecolor{red}{RGB}{255, 0, 0}
\definecolor{green}{RGB}{0, 100, 0}
\definecolor{gold}{RGB}{255, 125, 0}

\makeatletter
\DeclareRobustCommand\onedot{\futurelet\@let@token\@onedot}
\def\@onedot{\ifx\@let@token.\else.\null\fi\xspace}

\def\ie{\emph{i.e}\onedot}

\def\etc{\emph{etc}\onedot}

\def\etal{\emph{et al}\onedot}
\makeatother

\acrodef{nlp}[NLP]{natural language processing}
\acrodef{plm}[PLM]{pretrained language model}
\acrodef{sota}[SOTA]{state-of-the-art}
\acrodef{bs}[BS]{Beam Search}
\acrodef{mhs}[MHS]{Metropolis-Hastings Sampling}
\acrodef{hs}[HS]{Hybrid Search}
\acrodef{uas}[UAS]{unlabeled attachment score}
\acrodef{dda}[DDA]{Directed Dependency Accuracy}
\acrodef{sota}[SOTA]{state-of-the-art}
\acrodef{pos}[POS]{part-of-speech}
\acrodef{asr}[ASR]{attacking success rate}
\acrodef{ppl}[PPL]{Perplexity score}
\acrodef{cqr}[CQR]{Conversational Question Reformulation}
\acrodef{cqa}[CQA]{Conversational Question Answering}
\acrodef{mcqr}[MTCQR]{Multi-Topic Conversational Question Reformulation}
\acrodef{amt}[AMT]{Amazon Mechanical Turk}
\acrodef{mtcl}[MTCL]{Multi-Topic Contrastive Learning}

\def\vx{{\bm{x}}}
\def\vy{{\bm{y}}}

\title{Touch100k: A Large-Scale Touch-Language-Vision Dataset for Touch-Centric Multimodal Representation}

\author{% 
 Ning Cheng$^{1}$, Changhao Guan$^{1}$, Jing Gao$^{1}$, Weihao Wang$^{1}$, You Li$^{1}$, Fandong Meng$^{2}$, \\
 \textbf{Jie Zhou$^{2}$, Bin Fang$^{3}$, Jinan Xu$^{1}$, Wenjuan Han$^{1}$}  \\
 \\
 $^{1}$Beijing Jiaotong University, Beijing, China \quad
  $^{2}$WeChat AI, Tencent Inc., Beijing, China \\
  $^{3}$Beijing University of Posts and Telecommunications, Beijing, China
}
\begin{document}

\maketitle

\begin{abstract}
    Touch holds a pivotal position in enhancing the perceptual and interactive capabilities of both humans and robots. Despite its significance, current tactile research mainly focuses on visual and tactile modalities, overlooking the language domain. Inspired by this, we construct \dataset, a paired touch-language-vision dataset at the scale of 100k, featuring tactile sensation descriptions in multiple granularities (i.e., \textit{sentence-level} natural expressions with rich semantics, including contextual and dynamic relationships, and \textit{phrase-level} descriptions capturing the key features of tactile sensations). Based on the dataset, we propose a pre-training method, Touch-Language-Vision Representation Learning through Curriculum Linking (\model, for short), inspired by the concept of curriculum learning. \model aims to learn a tactile representation for the GelSight sensor and capture the relationship between tactile, language, and visual modalities. We evaluate our representation's performance across two task categories (namely, material property identification and robot grasping prediction), focusing on tactile representation and zero-shot touch understanding. 
    The experimental evaluation showcases the effectiveness of our representation. By enabling \model to achieve substantial improvements and establish a new state-of-the-art in touch-centric multimodal representation learning,  \dataset demonstrates its value as a valuable resource for research.
    Project page: \url{https://cocacola-lab.github.io/Touch100k/}.
\end{abstract}

\section{Introduction}\label{sec:intro}
Tactile perception constitutes a vital avenue through which humans interact with their surrounding environment. By touching objects, humans acquire information about the properties and structure of the objects, such as shapes, surface textures, temperatures, and hardness, among other characteristics. In the intricate realm of robotics, where robots are tasked with navigating complex environments and interacting with diverse objects, tactile perception enables precise manipulation and interaction. As the ability to comprehend the physical world through touch \cite{hansen2022visuotactile, qi2023general, zhou2023tactorbots}, tactile perception emerges as a crucial element in robotics, significantly advancing robots towards general purpose.

While research on tactile perception holds importance, it remains relatively underexplored compared to studies on visual and auditory perception. Current studies \cite{dave2024multimodal, li2019connecting, yang2022touch} mainly focus on integrating touch with vision to represent tactile sensations, with limited exploration in the realm of language modality. Despite many works \cite{gao2023objectfolder, kerr2022ssvtp, yang2022touch} involving language, the language presentation of these works is often in the form of textual classification labels. 

To fill this gap, we consider annotating paired vision-touch data with tactile sensation descriptions at multiple granularities. Specifically, we first collect and curate a total of 101,982 visual-tactile observations from publicly available tactile datasets as our foundational vision-touch dataset. Sequentially, we utilized GPT-4V \cite{achiam2023gpt}, along with carefully crafted prompts, to generate textual descriptions at multiple granularities. These descriptions contain rich tactile information. To ensure the accuracy and practicality of the descriptions, we further conduct multiple steps of quality enhancement, thus guaranteeing the data quality. As a result, we obtain 100,147 touch-language-vision data entries, while invalid data is manually filtered out. With this, we introduce a new dataset named Touch100k. To the best of our knowledge, Touch100k is the first dataset to encompass tactile, multi-granularity language, and visual modalities at a scale of 100k.

Based on the \dataset dataset, we propose a pretraining method, \textbf{T}ouch-\textbf{L}anguage-\textbf{V}ision Representation Learning through Curriculum \textbf{Link}ing (\textbf{\model}), to learn a tactile representation for the GelSight sensor \cite{yuan2017gelsight}. 
The linking process includes two stages: curriculum representation for tactile encoding and modality alignment. We adopt a teacher-student curriculum paradigm, where the well-established vision encoder acts as the teacher model, transferring knowledge to the student, our touch encoder. Specifically, the curriculum representation for tactile encoding is defined as a weighted combination of the visual representation and the tactile representation. 
Initially, the curriculum representation relies heavily on the teacher model due to the limited capacity of the student model. As pretraining progresses, the student model improves, allowing for a gradual decrease in the teacher's influence. Regarding the language modality, we encode multi-granularity textual descriptions using a text encoder and then fuse them to generate a final text feature. Subsequently, contrastive learning is employed to achieve alignment between the curriculum representation and the language modality. 
Finally, we evaluate the pretraining performance of \model on two types of tasks: \textit{material property identification} and  \textit{robot grasping prediction}. Experimental results demonstrate the effectiveness of the \dataset dataset and the advantages of the \model method.
In summary, the primary contributions of this paper are three-fold: data, pretraining method, and experiments. 
\begin{itemize}
\setlength{\itemsep}{0pt}
\setlength{\parsep}{0pt}
\setlength{\parskip}{0pt}
  \item We introduce \dataset (\S\ref{sec:data}), a large-scale paired touch-language-vision dataset to encompass tactile, multi-granularity language, and visual modalities, in touch scenarios.
  \item We present a touch-centric multimodal pretraining method (\S\ref{sec:method}), named Touch-Language-Vision
Representation Learning through Curriculum Linking (\model for short), to establish a tactile representation for the GelSight sensor.
  \item Experiments across various settings and tasks demonstrate the effectiveness of our dataset and pertaining method (\S\ref{sec:expt}).
\end{itemize}

\section{Related Work}
\label{sec:work}
\paragraph{Tactile Perception.} 
As one of the five fundamental human senses, tactile perception research is a key area for advancing towards universal robotics. The acquisition and application of tactile data play an indispensable role in driving research in tactile perception \cite{cui2020self, fazeli2019see, lin2019learning}. The collection of tactile data comes from tactile sensors. Present-day popular tactile sensors are vision-based tactile sensors, 
mainly including GelSight \cite{calandra2017feeling, dong2017improved, gomes2021generation, johnson2011microgeometry, li2019connecting, si2022taxim, yang2022touch, yuan2017shape}, DIGIT \cite{kerr2022self, lambeta2020digit, suresh2023midastouch}, and GelSlim \cite{gao2023objectfolder}. 
A major obstacle in tactile perception research is the considerable human, material, and financial resources required to construct high-quality tactile datasets. Thanks to the relentless efforts of researchers, there are some publicly available datasets: TVL \cite{fu2024touch}, ObjectFolder Real \cite{gao2023objectfolder}, SSVTP \cite{kerr2022self}, Touch and Go \cite{yang2022touch}, Objectfolder 2.0 \cite{gao2021objectfolder}, YCB-Slide \cite{suresh2022midastouch}, ObjectFolder 1.0 \cite{gao2021objectfolder}, VisGel \cite{li2019connecting}, ViTac Cloth \cite{luo2018vitac}, Data\_ICRA18 \cite{yuan2018active}, GelFabric \cite{yuan2017connecting}, the feeling of success \cite{calandra2017feeling}. However, these datasets provide a limited exploration of the language modality. They primarily focus on classification labels, hindering the potential for establishing richer cross-modal associations. 

\begin{table*}[t]
 \small
 \centering
 \caption{Comparison of the \dataset dataset to other tactile datasets. \textbf{TM.}: Tactile Modality. \textbf{VM.}: Visual Modality. \textbf{LM.}: Language Modality.  \textbf{ML.}: Multigranularity Language. \textbf{AS.}: Abundant Semantics. \textbf{MO.}: Multi-category Objects.  \textbf{DS.}: Data Size.} 
\label{tab:dataset_comparison}
% \begin{tabular}{|p{2 cm}|p{1 cm}|p{1 cm}|p{1 cm}|p{1 cm}|p{1 cm}|} \\ \hline
% \begin{tabular}{|c|c|p{1.8cm}|c|p{2.2 cm}|p{2cm}|}
\begin{tabular}
{c|c|c|c|c|c|c|c}
%{|c|c|p{2cm}|c|p{2.3cm}|p{2cm}|p{2 cm}|}
\toprule
\multirow{2}{*}{\textbf{Dataset}} &  
\multirow{2}{*}{\parbox{1.1cm}{\centering \textbf{TM.}}} & \multirow{2}{*}{\parbox{1.1cm}{\centering \textbf{VM.}}} &
\multirow{2}{*}{\parbox{1.1cm}{\centering \textbf{LM.}}}  & 
\multirow{2}{*}{\parbox{1.1cm}{\centering \textbf{MG.}}} &
\multirow{2}{*}{\parbox{1.1cm}{\centering \textbf{AS.}}} & \multirow{2}{*}{\parbox{1.1cm}{\centering \textbf{MO.}}} & 
\multirow{2}{*}{\parbox{1.1cm}{\centering \textbf{$\geq$100k DS.}}} \\
&&&&&&\\\midrule
\textbf{\dataset} & \cmark&\cmark &   \cmark  & \cmark & \cmark & \cmark & \cmark \\%Web and Wikipedia\\
\midrule
TLV~\cite{cheng2024towards} &\cmark &\cmark & \cmark  & \xmark & \cmark & \cmark & \xmark \ \\
TVL~\cite{fu2024touch} &\cmark &\cmark & \cmark  & \xmark & \cmark & \cmark & \xmark \ \\
OF Real~\cite{gao2023objectfolder} &\cmark &\cmark & \cmark  & \xmark & \xmark & \cmark & \cmark \ \\
SSVTP~\cite{kerr2022ssvtp} &\cmark &\cmark & \cmark  & \xmark & \xmark & \xmark & \xmark \ \\
TAG~\cite{yang2022touch} &\cmark &\cmark & \cmark & \xmark & \xmark & \cmark & \cmark \ \\
OF 2.0~\cite{gao2022objectfolder} &\cmark &\cmark & \cmark & \xmark & \xmark & \cmark & \cmark \ \\
YCB-Slide~\cite{suresh2022midastouch} &\cmark &\cmark & \cmark & \xmark & \xmark & \cmark & \cmark \ \\
OF 1.0~\cite{gao2021objectfolder} &\cmark &\cmark & \cmark & \xmark & \xmark & \cmark & \xmark \ \\
VisGel~\cite{li2019connecting} &\cmark &\cmark & \xmark & \xmark & \xmark & \cmark & \cmark \ \\
ViTac Cloth~\cite{luo2018vitac} &\cmark &\cmark & \xmark & \xmark & \xmark & \xmark & \xmark \ \\
Data\_ICRA18~\cite{yuan2018active} &\cmark &\cmark & \cmark & \xmark & \xmark & \xmark & \xmark \ \\
GelFabric ~\cite{yuan2017connecting} &\cmark &\cmark & \xmark & \xmark & \xmark & \xmark &  \xmark \ \\
Feel ~\cite{calandra2017feeling} &\cmark &\cmark & \cmark & \xmark & \xmark & \cmark & \xmark \ \\
\bottomrule 
\end{tabular}
%\vspace{-2mm}

\end{table*}

\paragraph{Multimodal Alignment.} 
In the real world, information is presented in multiple forms. For example, a picture may be accompanied by captions, and a video may contain multiple modalities such as speech and text. Multimodal alignment can help machines better understand the cross-modal information, enabling more effective processing and application. Radford \etal \cite{radford2021learning} established a bridge between visual and language modalities through self-supervised contrastive training.
Subsequent studies \cite{alayrac2022flamingo, jia2021scaling, yang2024empowering, zhao2024mmicl} further enhanced the alignment between visual and language modalities. With the advancement of technology, more modalities have been aligned, including audio, depth, thermal \etc. Girdhar \etal \cite{girdhar2023imagebind} broadened the shared representation space to six distinct modalities through image-centric contrastive learning, thereby enhancing a comprehensive understanding across modalities. Inspired by this approach, Zhu \etal \cite{zhu2023languagebind} proposed a language-centric framework, demonstrating its extensibility to any modality. They provided experiments with four modalities, all showing notable performance enhancements.

\paragraph{Curriculum Learning.}
Curriculum Learning (CL) is a machine learning training strategy that is similar to the structured progression of a human education curriculum. This approach has been proven to enhance the model's generalization capabilities \cite{soviany2022curriculum, DBLP:journals/corr/abs-2010-13166, wang2021survey}. The core concept of CL was first introduced by Bengio et al. \cite{bengio2009curriculum} to progressively raise the challenge level of the data fed to the model. 
Their initial investigations underscored the importance of commencing with less challenging data to facilitate smoother learning progression.
Hacohen and Weinshall \cite{hacohen2019power} contributed to the field by investigating how CL affects the training of deep neural networks. Their findings emphasized the benefits of CL in enhancing the efficiency and potency of deep learning models.
Wang et al. \cite{DBLP:journals/corr/abs-2010-13166} conducted a comprehensive survey on CL, providing a detailed analysis of its underlying principles, definitions, theories, and real-world applications. 
However, no prior work has explored its application to the task of multimodal alignment beyond vision and language.

\section{\dataset Dataset}\label{sec:data}
\begin{figure*}[t]
    %\vspace{-0.5em}
    \small
        \centering       
        \includegraphics[width=1.0\linewidth]{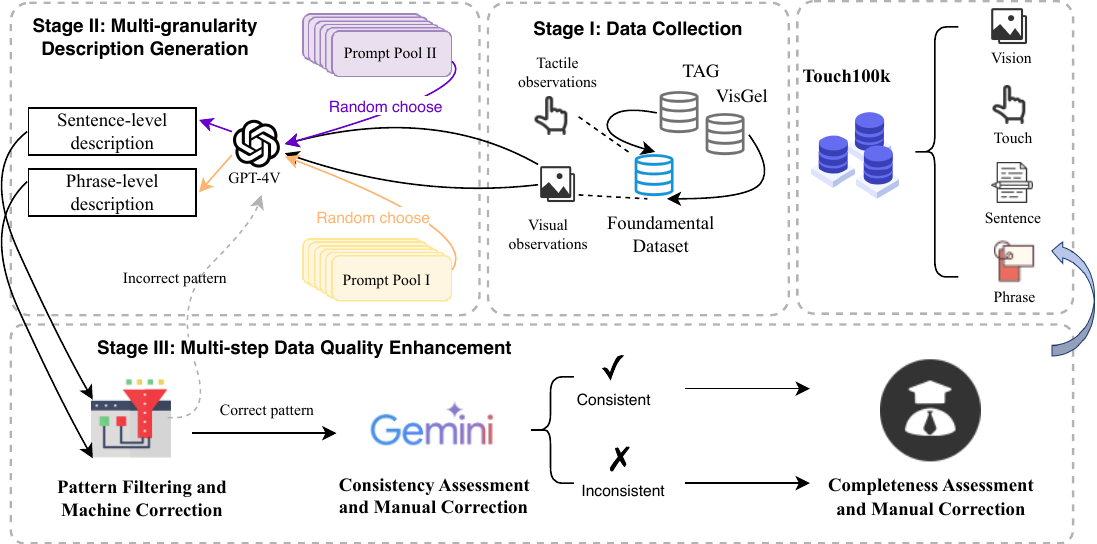}

        \caption{Construction process of the \dataset dataset.}       
        \label{fig:data_flow}
        %\vspace{-5mm}
    \end{figure*}

The \dataset dataset contains paired vision-and-touch data annotated with tactile sensations described in multi-granularity language. Table \ref{tab:dataset_comparison} shows \dataset shares many positive characteristics of existing tactile datasets while extending the granularity of descriptions and achieving significant data scaling-up. The construction process of \dataset is illustrated in Figure \ref{fig:data_flow}.

\subsection{Stage I: Data Collection}
\label{sec:data_s_1}
Our data sources are TAG \cite{yang2022touch} and VisGel \cite{li2019connecting} datasets. The former comprises a substantial volume of vision-touch data collected using the GelSight sensor, accompanied by classification labels. The latter also includes vision-touch data collected with GelSight but lacks textual labels. Hence, we make a point to manually label the names of the touched objects for the VisGel dataset. In total, we meticulously curated 101,982 touched observations,
with 91,982 from TAG and 10,000 from VisGel, to form the foundational vision-touch dataset for constructing \dataset.

\subsection{Stage II: Multi-granularity Description Generation}
\label{sec:data_s_2}
We utilize GPT-4V \footnote{gpt-4-turbo, the April and May 2024 versions.} \cite{achiam2023gpt} to generate multi-granularity textual descriptions. Visual and tactile observations offer distinct perspectives on the same semantic content. Leveraging this, we meticulously designed prompts and employed visual images as inputs to generate descriptions incorporating tactile sensations. 
The generated descriptions are multi-granular and consist of two levels: sentence-level and phrase-level. Sentence-level descriptions are natural expressions, offering a high-level perspective with rich semantics, including contextual relationships, containing dynamic relationships, and providing abundant information for tactile representation. Phrase-level descriptions capture the key features of tactile sensations. 
We individually crafted a prompt pool to generate descriptions for different granularities. 
One prompt was randomly selected from the prompt pool each time to ensure that the model captures subtle differences, thereby enhancing the diversity and depth of content. The prompt pools can be found in Appendix \ref{sec:appendix_prompt}.

\subsection{Stage III: Multi-step Data Quality Enhancement}
\label{sec:data_s_3}
To improve the reliability of the generated descriptions, we deploy a multi-step data quality enhancement process. This process is divided into three steps: pattern filtering and machine correction, consistency assessment and manual correction, and completeness assessment and manual correction.

\paragraph{Pattern Filtering and Machine Correction.} This step involves identifying and filtering out descriptions with erroneous patterns using regex-based pattern matching, targeting issues like multilanguage confusion, special markers, or semantic redundancy. These screened problematic data will be re-entered into GPT-4V for correction.

\paragraph{Consistency Assessment and Manual Correction.} We use Gemini \footnote{Gemini 1.0 Pro.} \cite{team2023gemini}, another powerful multimodal model, as a referee to assess the consistency of tactile sensations between the input visual image and the descriptions generated by GPT-4V. Through this assessment, the workers we hired can effectively identify and correct descriptions that are inconsistent with the image, further enhancing the quality of the descriptions.

\paragraph{Completeness Assessment and Manual Correction.} Through the two steps mentioned above, we found that some generated descriptions lack completeness in terms of tactile sensations. Therefore, we conducted human evaluations to ensure that the generated descriptions convey various aspects of tactile sensations, including touch point locations, texture features, and so on. This is crucial to alignment between tactile and language modalities, especially when applying the aligned tactile representation to various downstream tasks. Any deficiencies we found were further refined and supplemented manually.

\begin{figure}[t]
    %\vspace{-0.5em}
    \small
        \centering       
        \includegraphics[width=1.0\linewidth]{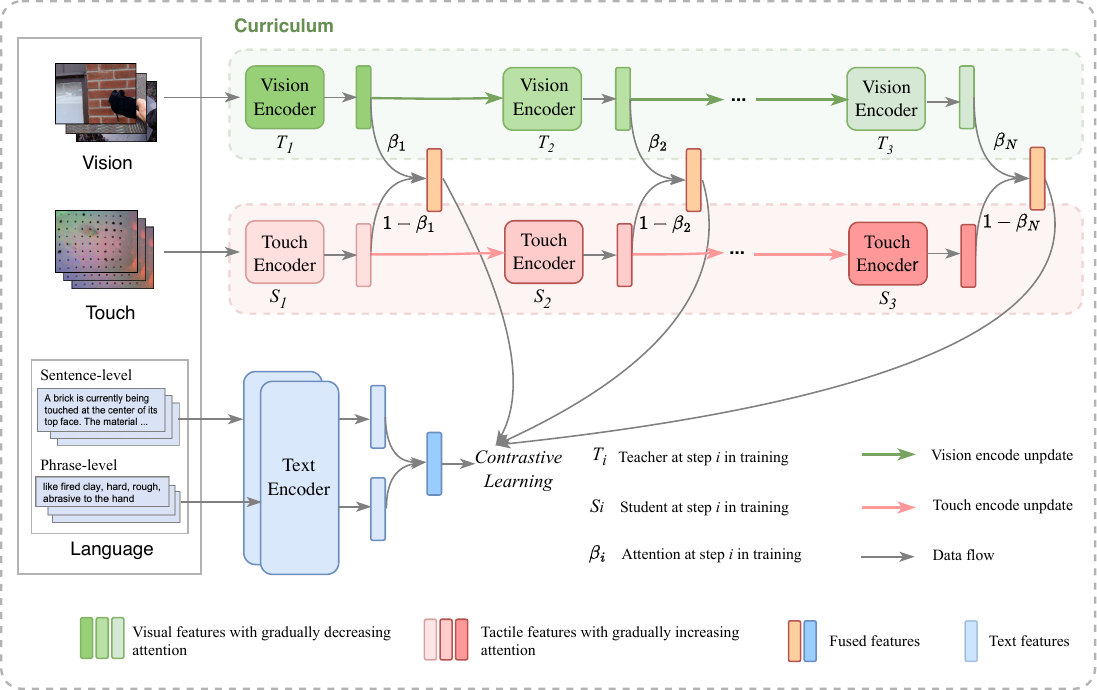}

        \caption{\model overview. The text encoder parameters are frozen, while the parameters of the vision and touch encoders can be adjusted. Since the parameters of the text encoder are frozen, sentence-level descriptions and phrase-level descriptions are sequentially fed into a single text encoder. For illustrative purposes, two text encoders are depicted in this figure.}       
        \label{fig:method}
        %\vspace{-5mm}
    \end{figure}

\section{Method}\label{sec:method}
We link the tactile, language, and visual modalities aiming to learn a tactile representation for the GelSight sensor. This leads to our method, called \textbf{T}ouch-\textbf{L}anguage-\textbf{V}ision Representation Learning through Curriculum \textbf{Link}ing (\model, for short). Figure \ref{fig:method} presents an overview of \model. 
Linking three modalities follows a two-stage protocol: curriculum representation and modality alignment.
We first introduce curriculum representation (\S\ref{subsec:curri}), which is inspired by the concept of curriculum learning for tactile encoding, and then introduce modality alignment (\S\ref{subsec:align}), following the principles of contrastive learning.

\subsection{Curriculum Representation for Tactile Encoding}
\label{subsec:curri}
Leveraging the inherent semantic overlap between visual and tactile modalities, this work exploits the well-established visual understanding capabilities of pre-trained visual encoders. We adopt a teacher-student curriculum training paradigm (Figure \ref{fig:method}), where the vision encoder serves as the teacher model and the touch encoder acts as the student model. During training, the teacher model transfers knowledge to the student model.

Formally, we define the curriculum representation $\vx$ as a weighted combination of the visual representation, $x^{(v)}$, and the tactile representation, $x^{(t)}$, using a pre-defined weight $\beta_i \in [0, 1]$ as follows:

\begin{equation}
\label{eq:lora}
\vx = \beta_{i} {x^{(v)}} + (1-\beta_{i}) {x^{(t)}}
\end{equation}

This facilitates knowledge transfer by leveraging the strong visual representation as a guide for the student model to learn tactile representations.

Furthermore, to enhance knowledge transfer, we incorporate curriculum learning. Initially, the student model's limited capacity necessitates a heavier reliance on the teacher model, reflected by a larger initial weight $\beta_1$. As training progresses, the student model progressively improves, allowing for a gradual decrease in the teacher's influence (decreasing $\beta_i$) until the student model matures.
Formally, at the $i$-th training step, $\beta_i$ undergoes linear decay and can be defined as
\begin{equation}
\label{eq:decay}
\beta_{i} = \beta_{1} - \frac{i}{N} (\beta_{1}-\beta_{min})
\end{equation}
where $N$ is the total number of training steps, and $\beta_{min}$ denotes ultimate weight. Here we set the value of $\beta_{min}$ to 0.

\subsection{Modality Alignment}
\label{subsec:align}
We align the curriculum representation for tactile encoding $\vx$ with the language modality to assist in tactile-related downstream tasks, such as material property identification and grasp stability prediction. Specifically, multi-granularity textual descriptions are encoded by the text encoder to derive their respective representations, which are then fused to yield the final text feature $\vy$. We leverage the text encoder from OpenCLIP-large \cite{ilharco_gabriel_2021_5143773}. Benefiting from its pretraining on large-scale datasets and strong language representation capabilities, we keep the text encoder frozen without parameter updating during training. We perform contrastive learning \cite{radford2021learning} for the alignment of touch and language modalities. The touch encoder (full training) and vision encoder (LoRA \cite{hu2021lora} fine-tuning) are optimized using an InfoNCE loss:
\begin{equation}\label{eq:loss}
\begin{aligned}
L=-\frac{1}{K} \sum_{i=1}^{K} \log \frac{\exp (\vx_i^{\top} \vy_i / \tau)}{\sum_{j=1}^{K} \exp (\vx_i^{\top} \vy_j / \tau)}    \\
\end{aligned}
\end{equation}
where $\tau$ denote the scalar factor, and $K$ is the batch size.

\begin{table}
%\small
%\vspace{2mm}
\centering
\caption{Accuracy of different models on \textit{material property identification} and  \textit{robot grasping prediction} tasks. The best performance is \textbf{bold}, and the suboptimal is \underline{underlined}. The best with performance improvement compared to the suboptimal within 0.5\% are marked as \textcolor{green}{green} and above 0.5\% are highlighted in \textcolor{myred}{red}.} 
\label{tab:method-results}
\resizebox{399pt}{!}{
\begin{tabular}{llcccc}
\toprule
\multirow{2}{*}{\textbf{Model}} & \multirow{2}{*}{\textbf{Training Data}} & \multicolumn{3}{c}{\textbf{Material Property}}  & \multirow{2}{*}{\textbf{Robot Grasping}} \\ 
\cmidrule{3-5}
& & Material & Hard/Soft & Rough/Smooth & \\
%\cmidrule{2-5}
\midrule
 Chance & - & 5.0 & 50.0 & 50.0 & 50.0 \\ 
 \noalign{\vskip 0.4ex} \cdashline{1-6} \noalign{\vskip 0.6ex}
 \multicolumn{4}{l}{\textit{Linear Probing}} \\
Supervised & ImageNet & 46.9  &  72.3  & 76.3  & 73.0       \\
VT CMC \cite{yang2022touch} & TAG & 54.7  &  77.3  & 79.4  & 78.1       \\
MViTac \cite{dave2024multimodal} & TAG / Feel & 57.6  &  86.2  & 82.1  & -      \\
UniTouch \cite{yang2024binding}  & TAG, Feel, YCB-Slide, OF 2.0 & 61.3  & - & -  & \underline{82.3}      \\
VIT-LENS-2 \cite{lei2023vit} & TAG & \underline{63.0}  &  \underline{92.0}  & \textbf{85.1} (\textcolor{green}{+0.4})  & -   \\
\model (Ours) & Touch100k  & \textbf{67.2} (\textcolor{red}{+4.2}) & \textbf{93.1} (\textcolor{red}{+1.1})  & \underline{84.7}  & \textbf{94.5} (\textcolor{red}{+12.2}) \\
\midrule
\multicolumn{4}{l}{\textit{Zero-Shot}} \\
UniTouch \cite{yang2024binding}  & TAG, Feel, YCB-Slide, OF 2.0 & 52.7  & - & -  & \textbf{65.5} (\textcolor{green}{+0.1})      \\
VIT-LENS-2 \cite{lei2023vit} & TAG & \underline{65.8}  &  \underline{74.7}  &  \underline{63.8}  & -       \\
\model (Ours) & Touch100k  & \textbf{70.0} (\textcolor{red}{+4.2})  & \textbf{79.3} (\textcolor{red}{+4.6})  & \textbf{77.8} (\textcolor{red}{+14.0})   & \underline{65.4} \\
\bottomrule
\end{tabular}
}
%\vspace{-3mm}
\end{table}

\section{Experiments}\label{sec:expt}
We train our representation model, \model, on \dataset dataset, and evaluate the representation's performance on two types of tasks: \textit{material property identification} and  \textit{robot grasping prediction}.
For \textit{material property identification}, we conduct experiments on the three test sets of TAG \cite{yang2022touch}. These three test sets correspond to three classification subtasks: (1) material classification, (2) hard/soft classification, and (3) rough/smooth classification. Among them, the first is a multi-class classification task with 20 object labels, and the other two are binary classification tasks. For \textit{robot grasping prediction}, we use the Feel \cite{calandra2017feeling} dataset for evaluation. The goal of this task is to predict whether a robotic gripper can successfully grasp an object between its left and right fingers based on 4 tactile observations (before/after observations for left and right tactile sensors). Since the Feel dataset lacks a unified split into train/valid/test sets \cite{yang2022touch, gao2023objectfolder, yang2024binding}, following \cite{yang2024binding}, we split the dataset by objects in the ratio of 8:1:1.

\subsection{Implementation Details}
For visual and tactile modalities, we use the 24-layer, 1024-dimensional Vision Transformer (ViT) \cite{dosovitskiy2020image} with a patch size of 14. Both the vision encoder and the touch encoder are initialized from OpenCLIP-large \cite{ilharco_gabriel_2021_5143773}. Tactile observations are treated as RGB images. We use the text encoder from OpenCLIP-large for textual descriptions of different granularities. We opt for full training solely for the touch encoder. Given the robust textual and visual representation capabilities of the visual and text encoders from OpenCLIP-large, we keep the text encoders frozen, and employ LoRA fine-tuning for the visual encoder to adapt it to visual observations in tactile scenarios. While the training process involves multiple encoders, our focus on adapting the visual encoder to the touch domain motivates our terminology: touch-centric multimodal representation learning.

\subsection{Comparative Models}
We use random classification (\ie, Chance) and supervised ImageNet \cite{deng2009imagenet} features as a reference. We then compare our model's performance against a series of recent state-of-the-art models: VT CMC \cite{yang2022touch}, MViTac \cite{dave2024multimodal}, UniTouch \cite{yang2024binding}, and VIT-LENS-2 \cite{lei2023vit}. VT CMC and MViTac only rely on vision and touch encoders for training. Therefore, when performing two types of tasks, these models require linear probing instead of direct zero-shot evaluation. In contrast, benefiting from the utilization of the language model, UniTouch and VIT-LEN-2 possess zero-shot capability. Among them, MViTac requires training separate tactile encoders for each task, while other methods employ a single shared touch encoder.

\subsection{Results}
We conducted two types of settings: linear probing and zero-shot. Linear probing provides an assessment of the tactile encoder, measuring the performance of tactile representations on specific tasks, while zero-shot evaluation rigorously evaluates the understanding and generalization capabilities of the touch encoder. Following previous work \cite{dave2024multimodal, gao2023objectfolder, yang2022touch, yang2024binding}, we use accuracy as the evaluation metric.

\paragraph{Tactile representation.}
To evaluate the quality of the learned tactile representations, we conduct experiments within a linear probing setting. This setup involves training a separate, lightweight linear classification head on top of the frozen touch encoder for specific downstream tasks.
The middle block of Table \ref{tab:method-results} represents the results of linear probing. Except for a slight bias in the rough/smooth classification subtask, our method generally outperforms all state-of-the-art comparison models in two types of tasks (\ie, \textit{material property identification} and \textit{robot grasping prediction}), especially in the robot grasping task.

\paragraph{Zero-shot touch understanding.}
We leverage zero-shot evaluation to assess our model's ability to understand tactile information. As shown in the bottom block of Table \ref{tab:method-results}, our method, \model, achieves superior performance on both task types. This finding highlights the strong perception and generalization capabilities of our touch encoder, enabling it to effectively handle unseen data distributions.

\begin{table}[t]
\centering
\caption{Comparison to TAG dataset on downstream tasks. TTCL.: Touch-Text Contrastive Learning.}
\label{tab:dataset-results}
\resizebox{399pt}{!}{
\begin{tabular}
{lccccc}
%{|c|c|p{2cm}|c|p{2.3cm}|p{2cm}|p{2 cm}|}
\toprule
\multirow{2}{*}{ \textbf{\small Dataset}} & \multirow{2}{*}{\textbf{\small Method}} & \multicolumn{3}{c}{\textbf{\small Material Property}}  & \multirow{2}{*}{\textbf{\small Robot Grasping}} \\ 
\cmidrule{3-5}
& & \small Material & \small Hard/Soft & \small Rough/Smooth & \\
\midrule
\small Chance & \small - & \small 18.6 & \small 66.1 & \small 56.3 & \small 56.1 \\
 \noalign{\vskip 0.4ex} \cdashline{1-6} \noalign{\vskip 0.6ex}
\small TAG & \multirow{2}{*}{\small VT CMC + TTCL. } & \small 60.8 & \small 76.6 & \small 46.9 & \small  43.4 \\
\small Touch100k (Ours) & \small  & \small \textbf{64.1} & \small \textbf{79.3} & \small \textbf{72.0} & \small \textbf{61.3} \\
\bottomrule 
\end{tabular}
}
%\vspace{-3mm}
\end{table}

\begin{table}[t]
\centering
\caption{Impact of dataset scale on downstream performance. }
\label{tab:dataset-scale}
\resizebox{399pt}{!}{
\begin{tabular}
{lccccc}
%{|c|c|p{2cm}|c|p{2.3cm}|p{2cm}|p{2 cm}|}
\toprule
\multirow{2}{*}{\textbf{\small Dataset}} & \multirow{2}{*}{\textbf{\small Scale }} & \multicolumn{2}{c}{\textbf{\small Material Classification}}  & \multicolumn{2}{c}{\textbf{\small Robot Grasping}} \\ 
\cmidrule{3-6}
& & \textit{\small Linear Probing} & \textit{\small Zero-Shot} & \textit{\small Linear Probing} & \textit{\small Zero-Shot} \\
\midrule

\multirow{4}{*}{\small Touch100k}
& \small 25k - 25\% of the total & \small 62.3 & \small 26.5 & \small 92.6 & \small 44.7 \\
& \small 50k - 50\% of the total &\small 64.7 & \small 61.7 & \small 92.3 & \small 50.6 \\
& \small 75k - 75\% of the total & \small 66.4 & \small 66.7 & \small 93.1 & \small 52.1 \\
& \small 100k - 100\% of the total& \small \textbf{67.2} & \small \textbf{70.0} & \small \textbf{94.5} & \small \textbf{65.4} \\
\bottomrule 
\end{tabular}
}
%\vspace{-3mm}
\end{table}

\subsection{Dataset Analysis}

\paragraph{Comparison to TAG dataset.}
To evaluate the effectiveness of our dataset, we compared our \dataset with the TAG dataset using the same method, \ie, VT CMC + TTCL. The method is our refinement of VT CMC to suit scenarios involving textual descriptions. Specifically, we divide the training process into two stages. In the first stage, we follow the experimental setup of VT CMC, conducting joint training with tactile and visual modalities. Then, in the second stage, contrastive learning is employed for the touch encoder from the first stage along with a text encoder from OpenCLIP-large.
This text encoder maintained consistency with the settings of the aforementioned experiments and remained frozen during training. Notably, the CMC features of the TAG dataset outperform those of VisGel and ObjectFolder 2.0. Consequently, we consider the TAG dataset features as the upper bound for VisGel and ObjectFolder 2.0, and thus, we opted to exclude comparisons with non-state-of-the-art baselines in Table~\ref{tab:dataset-results} for brevity. From Table \ref{tab:dataset-results}, it can be observed that our dataset provides a useful signal for training tactile representations.

\paragraph{Impact of dataset scale.}
We explored the impact of dataset scale on downstream performance. We respectively took 75\%, 50\%, and 25\% of the total dataset size to obtain three datasets of different scales, with data volumes corresponding to 75k, 50k, and 25k samples. Table \ref{tab:dataset-scale} shows the experimental results. We are surprised to find that reducing the data scale has little impact on the linear probing setting. Taken together, the results demonstrate that increasing the dataset scale can positively influence experimental outcomes, leading to better downstream performance. Simultaneously, zero-shot experiments indicate that our pretraining method can comprehensively learn features and patterns within the data, thus enhancing its generalization ability. This suggests that our tactile representation captures the underlying structure of the data on a larger scale, thereby performing well on unseen data.

\begin{figure*}[t]
    \centering
    \includegraphics[width=\linewidth]{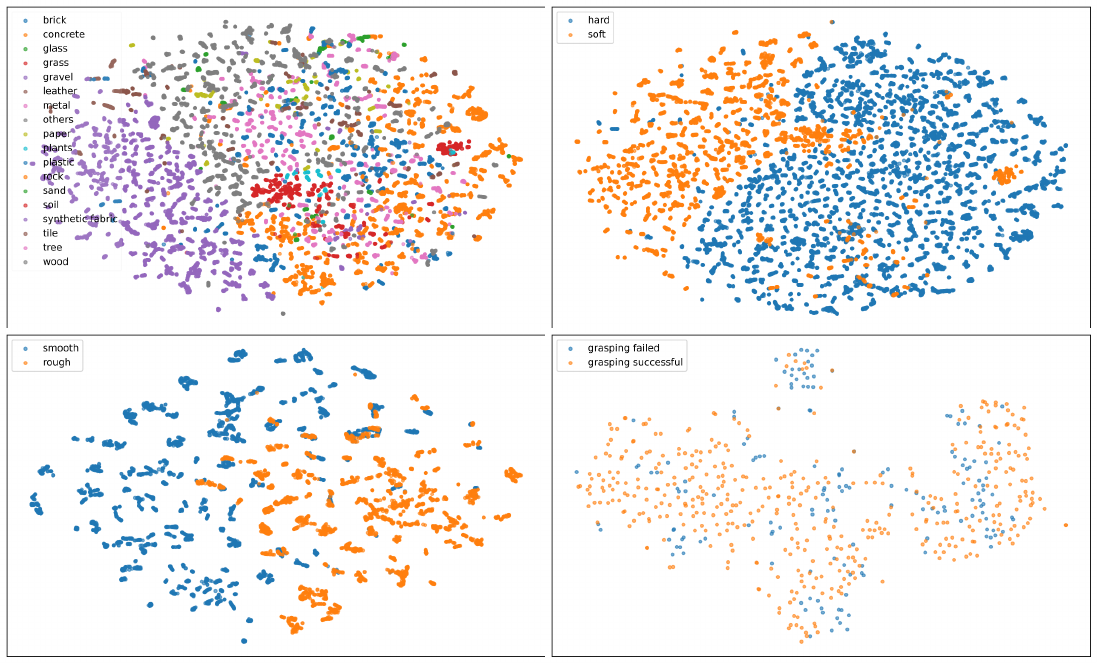}
    \caption{t-SNE projection on various subtasks.}
    \label{fig:case_study}
    %\vspace{-5mm}
\end{figure*}

\subsection{Tactile Representation Analysis}
We adopt t-SNE~\cite{van2008visualizing} to project learned tactile representation into a 2-dimensional space. From the visualization shown in Figure \ref{fig:case_study}, we can see our pretraining method, \model, excels at Hard/Soft and Rough/Smooth classification. On both of these subtasks, \model effectively separates the two categories. However, when it comes to tasks involving multi-classification and significantly different data distributions, the generalization ability of its representations tends to decrease. This highlights the limitations of current approaches in tactile multi-classification and robot manipulation tasks and emphasizes the need for further development to achieve robustness.

\subsection{Ablation Study}
To further investigate the contribution of curriculum representation to \model, we conduct ablation studies by removing this module and comparing the results on four sub-tasks. The overall setups are divided into linear probing and zero-shot, with all other configurations remaining consistent, except for the removal of the module under investigation. The results are shown in Table \ref{tab:ablation}. When the curriculum representation is removed, we observed a performance drop across all tasks in both linear probing and zero-shot setups. The decline is pronounced in the robot grasping task, especially in the zero-shot evaluation, where performance dropped by 21.2\% (65.4\% \textit{vs.} 44.2\%). The success of our curriculum representation approach strongly supports the notion that curriculum learning can generalize effectively to new tasks, as evidenced by prior research \cite{soviany2022curriculum, DBLP:journals/corr/abs-2010-13166, wang2021survey}-- curriculum learning can generalize well to other tasks.

\begin{table}[t]
\small
%\vspace{-2mm}
\centering
\caption{Ablation study on material property identification and robot grasping prediction in linear probing and zero-shot setting. The percentage decrease in accuracy is marked in \textcolor{green}{green}.} 
\label{tab:ablation}
\resizebox{399pt}{!}{
\begin{tabular}{lcccc}
\toprule
\multirow{2}{*}{\textbf{Model}} & \multicolumn{3}{c}{\textbf{Material Property}} & \multirow{2}{*}{\textbf{Robot Grasping}}  \\ 
\cmidrule{2-4}
& Material  & Hard/Soft    & Rough/Smooth  &          \\ 
\midrule
\multicolumn{4}{l}{\textit{Linear Probing}} \\
\textbf{\model}  &  \textbf{67.2}  & \textbf{93.1}  & \textbf{84.7} & \textbf{94.5}        \\
- w/o Multi-stage Curriculum Representation  &  66.4 (\textcolor{green}{-0.8}) & 92.5 (\textcolor{green}{-0.6})  &  84.4 (\textcolor{green}{-0.3})  & 91.8 (\textcolor{green}{-2.7}) \\
\midrule
\multicolumn{4}{l}{\textit{Zero-Shot}} \\
\textbf{\model}  &  \textbf{70.0}  & \textbf{79.3}  & \textbf{77.8} & \textbf{65.4}        \\
- w/o Multi-stage Curriculum Representation  &  68.7 (\textcolor{green}{-1.3})  & 78.5 (\textcolor{green}{-0.8}) &  76.8 (\textcolor{green}{-1.0}) & 44.2 (\textcolor{green}{-21.2})\\
\bottomrule
\end{tabular}
}
%\vspace{-5mm}
\end{table}

\section{Conclusion and Future Work}\label{sec:conclusion}
In this work, we construct the first paired touch-language-vision dataset featuring tactile sensation descriptions in multiple granularities and name the \dataset dataset based on its size. We also propose a pertaining method (\model) for \dataset to obtain a tactile representation suitable for GelSight sensors. Experiments demonstrate the effectiveness of our dataset. 

Since our dataset and pertaining method are specifically designed for GelSight sensors, their generalizability to other tactile sensors, such as DIGIT or GelSlim, remains an open question. Future work will investigate techniques to enhance the transferability of tactile representations across diverse sensor types.

%%%%%%%%%%%%%%%%%%%%%%%%%%%%%%%%%%%%%%%%%%%%%%%%%%%%%%%%%%%%
\newpage
{%\small
    \bibliographystyle{plain}
    \bibliography{custom}
}

%%%%%%%%%%%%%%%%%%%%%%%%%%%%%%%%%%%%%%%%%%%%%%%%%%%%%%%%%%%%
\newpage
\section*{Checklist}

% %%% BEGIN INSTRUCTIONS %%%
% The checklist follows the references.  Please
% read the checklist guidelines carefully for information on how to answer these
% questions.  For each question, change the default \answerTODO{} to \answerYes{},
% \answerNo{}, or \answerNA{}.  You are strongly encouraged to include a {\bf
% justification to your answer}, either by referencing the appropriate section of
% your paper or providing a brief inline description.  For example:
% \begin{itemize}
%   \item Did you include the license to the code and datasets? \answerYes{See Section~\ref{gen_inst}.}
%   \item Did you include the license to the code and datasets? \answerNo{The code and the data are proprietary.}
%   \item Did you include the license to the code and datasets? \answerNA{}
% \end{itemize}
% Please do not modify the questions and only use the provided macros for your
% answers.  Note that the Checklist section does not count towards the page
% limit.  In your paper, please delete this instructions block and only keep the
% Checklist section heading above along with the questions/answers below.
% %%% END INSTRUCTIONS %%%

\begin{enumerate}

\item For all authors...
\begin{enumerate}
  \item Do the main claims made in the abstract and introduction accurately reflect the paper's contributions and scope?
    \answerYes{See the abstract and introduction part.}
  \item Did you describe the limitations of your work?
    \answerYes{See Section \ref{sec:conclusion} and Appendix \ref{sec:limitations}.}
  \item Did you discuss any potential negative societal impacts of your work?
    \answerNo{We focus on touch-centric multimodal research. We discuss the positive impacts of this research in Section \ref{sec:intro} and Section \ref{sec:work}. At present, we have not identified any negative societal impacts resulting from this foundational research.}
  \item Have you read the ethics review guidelines and ensured that your paper conforms to them?
    \answerYes{We have checked the ethics review guidelines, and our research is in line with them.}
\end{enumerate}

\item If you are including theoretical results...
\begin{enumerate}
  \item Did you state the full set of assumptions of all theoretical results?
    \answerNA{}
	\item Did you include complete proofs of all theoretical results?
    \answerNA{}
\end{enumerate}

\item If you ran experiments (e.g. for benchmarks)...
\begin{enumerate}
  \item Did you include the code, data, and instructions needed to reproduce the main experimental results (either in the supplemental material or as a URL)?
    \answerYes{See our \href{https://dingmyu.github.io/physion_v2/}{project page} or the supplemental material.}
  \item Did you specify all the training details (e.g., data splits, hyperparameters, how they were chosen)?
    \answerYes{All our training and testing details can be found in Section \ref{sec:expt} and Appendix \ref{sec:appendix_hyper}.}
	\item Did you report error bars (e.g., with respect to the random seed after running experiments multiple times)?
    \answerNo{We follow previous work \cite{dave2024multimodal, lei2023vit, yang2024binding, yang2022touch} and don't report error bars.}
	\item Did you include the total amount of compute and the type of resources used (e.g., type of GPUs, internal cluster, or cloud provider)?
    \answerYes{See Appendix \ref{sec:appendix_hyper}.}
\end{enumerate}

\item If you are using existing assets (e.g., code, data, models) or curating/releasing new assets...
\begin{enumerate}
  \item If your work uses existing assets, did you cite the creators?
    \answerYes{All assets we use are open source, and we have properly mentioned and cited them.}
  \item Did you mention the license of the assets?
    \answerYes{See the supplemental material.}
  \item Did you include any new assets either in the supplemental material or as a URL?
    \answerYes{See our \href{https://dingmyu.github.io/physion_v2/}{project page} or the supplemental material.}
  \item Did you discuss whether and how consent was obtained from people whose data you're using/curating?
    \answerYes{We claim that our work is based on publicly available data and provide appropriate citations.}
  \item Did you discuss whether the data you are using/curating contains personally identifiable information or offensive content?
    \answerYes{See the supplemental material.}
\end{enumerate}

\item If you used crowdsourcing or conducted research with human subjects...
\begin{enumerate}
  \item Did you include the full text of instructions given to participants and screenshots, if applicable?
    \answerYes{See the supplemental material.}
  \item Did you describe any potential participant risks, with links to Institutional Review Board (IRB) approvals, if applicable?
    \answerNA{There are no potential participant risks.}
  \item Did you include the estimated hourly wage paid to participants and the total amount spent on participant compensation?
    \answerYes{See the supplemental material.}
\end{enumerate}

\end{enumerate}

%%%%%%%%%%%%%%%%%%%%%%%%%%%%%%%%%%%%%%%%%%%%%%%%%%%%%%%%%%%%
\newpage
\appendix
\section{Data Statistical Analysis}
\begin{figure}[!ht]
    \centering
\includegraphics[width=\linewidth]{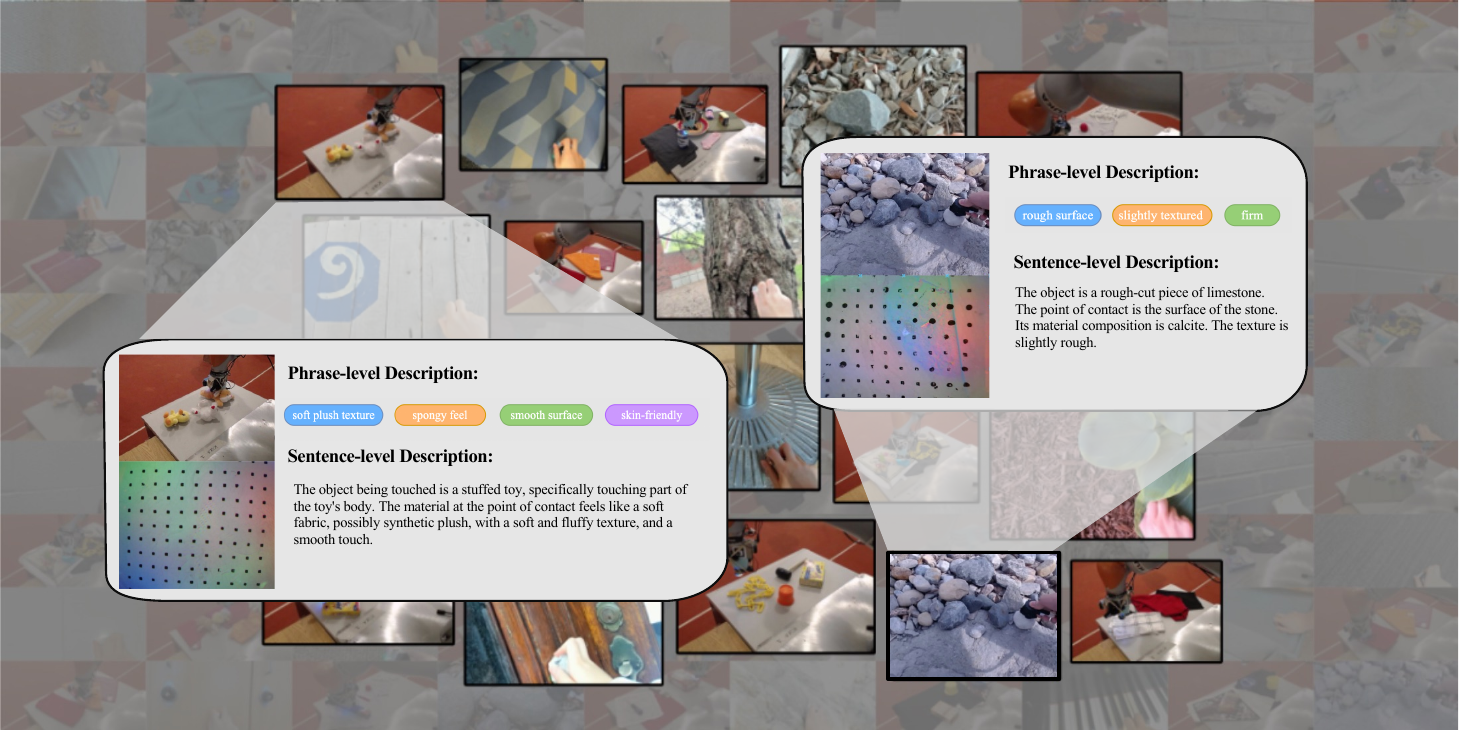}
    \caption{Illustration of examples in \dataset dataset.}
    \label{fig:task_example}
    \vspace{-3mm}
\end{figure}

\dataset contains 100,147 paired touch-language-vision samples. Several examples in \dataset are visualized in Figure~\ref{fig:task_example}. 

\begin{figure*}[!ht]
%\vspace{-5mm}
    \centering
    \includegraphics[width=\linewidth]{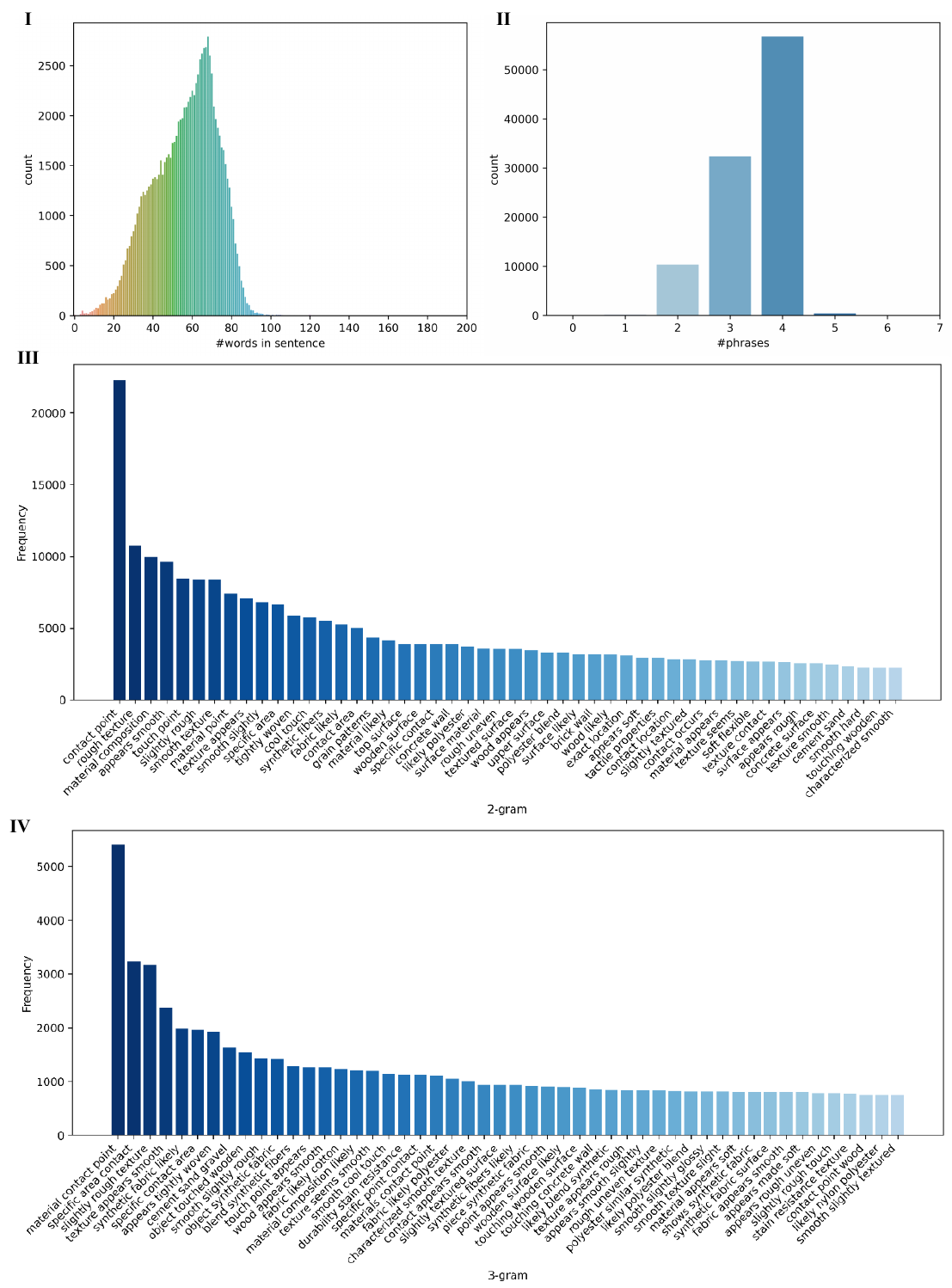}
    \caption{Statistical distributions on tactile sensation descriptions in multiple granularities (sentence-level descriptions: \textbf{I}, \textbf{III}, \textbf{IV}; phrase-level descriptions: \textbf{II}) for \dataset. \textbf{I}: The word count distribution in sentence-level descriptions. \textbf{II}: The phrase count distribution in phrase-level descriptions. \textbf{III} \& \textbf{IV}: The frequency distribution of the top 50 2-gram and 3-gram in sentence-level descriptions, showcasing contextual relationships. For \textbf{III} and \textbf{IV}, certain words in the sentence-level descriptions, lacking significant semantic value, are treated as stop words.}
    \label{fig:data_statistic}
\end{figure*}

We present the statistical analysis of tactile sensation descriptions in multiple granularities in Table~\ref{tab:dataset_static}, and the statistical distributions in Figure \ref{fig:data_statistic}. 

\begin{table}[!ht]
\vspace{-3mm}
\centering
\caption{Statistics on language at different granularities for \dataset. \textit{Avg. \#words}: The average number of words. \textit{Avg. \#phrases}: The average number of phrases. \textit{Avg. \#words in phrase}: The average number of words in a phrase.}
\label{tab:dataset_static}
%\resizebox{399pt}{!}{
\begin{tabular}
{p{2cm}p{3.3cm}p{3.3cm}p{3.3cm}}
%{|c|c|p{2cm}|c|p{2.3cm}|p{2cm}|p{2 cm}|}
\toprule
\multirow{2}{*}{\textbf{\small \makecell[c]{Dataset}}} & \multicolumn{1}{c}{\textbf{\small Sentence-level Descriptions}}  & \multicolumn{2}{c}{\textbf{\small Phrase-level Descriptions}} \\ 
\cmidrule{2-4}
& \textit{\small \makecell[c]{Avg. \#words}}  & \textit{\small \makecell[c]{Avg. \#phrases}} & \textit{\small \makecell[c]{Avg. \#words in phrase}} \\
\midrule

\multirow{1}{*}{\small \makecell[c]{Touch100k}}
& \small \makecell[c]{61} & \small \makecell[c]{4} & \small \makecell[c]{2}  \\
\bottomrule 
\end{tabular}
%}
\vspace{-2mm}
\end{table}

To provide further insights into the data, we have provided a list and counts of frequently occurring phrases (above 200 occurrences) as follows.

{'cool': 40551, 'rough': 35394, 'hard': 25524, 'smooth': 24181, 'soft': 16716, 'bumpy': 15764, 'rough texture': 10931, 'solid': 9464, 'scratchy': 8083, 'gritty': 7381, 'dry': 6515, 'cool to the touch': 6242, 'thin': 6149, 'slightly rough': 5863, 'coarse': 5387, 'warm': 5121, 'flexible': 5031, 'rough and coarse texture': 4390, 'rigid': 4307, 'cool temperature': 4057, 'cold': 3776, 'flat': 3642, 'gritty texture': 3510, 'splintery': 3269, 'unyielding': 2927, 'solid and unyielding': 2806, 'hard and unyielding touch': 2727, 'metallic': 2665, 'ticklish': 2625, 'firm resistance': 2350, 'grainy': 2267, 'slightly abrasive': 2222, 'rough and bumpy texture': 2205, 'grainy texture': 2027, 'distinct tactile sensation': 2009, 'cool feeling': 2001, 'rigid feel': 1982, 'fuzzy': 1948, 'light': 1941, 'hard and unyielding': 1894, 'fibrous': 1751, 'smooth surface': 1698, 'rough surface under fingertips': 1694, 'solid and sturdy': 1689, 'sharp edges': 1665, 'cool and solid': 1615, 'slightly damp': 1563, 'cool and solid surface': 1508, 'silky': 1502, 'hard surface': 1485, 'rough and scratchy': 1484, 'cool and smooth sensation': 1471, 'cool and smooth': 1469, 'no give': 1457, 'rough and gritty texture': 1448, 'stiff': 1423, 'cool surface': 1372, 'prickly': 1323, 'bumpy surface': 1291, 'cylindrical': 1237, 'slight indentations': 1234, 'smooth texture': 1210, 'slightly scratchy': 1206, 'delicate touch': 1166, 'plush': 1161, 'slightly fuzzy': 1124, 'heavy': 1108, 'slight give': 1070, 'rough and gritty': 1060, 'firm fingertip press': 1032, 'velvety': 1029, 'reflective': 1018, 'heavy skin brush': 934, 'rough and bumpy': 929, 'warm feeling': 906, 'slight friction': 855, 'synthetic': 852, 'delicate': 850, 'warm and inviting': 843, 'light skin brush': 817, 'strong tactile sensation': 811, 'uniform': 802, 'cool and hard': 791, 'coarse fingertip touch': 747, 'wooden': 742, 'slightly sharp': 737, 'loose': 722, 'cheap': 630, 'deep skin brush': 627, 'cool and smooth surface': 616, 'fluffy': 605, 'slightly rough texture': 602, 'springy': 594, 'inflexible': 589, 'slippery': 570, 'abrasive': 566, 'strong fingertip touch': 561, 'slightly tacky': 558, 'rough and bumpy surface': 558, 'gentle fingertip touch': 556, 'urban and modern feel': 553, 'rough and scratchy texture': 547, 'rough surface': 545, 'rough texture under fingers': 538, 'supple': 537, 'slightly gritty': 529, 'damp': 521, 'subtle tactile sensation': 505, 'slightly porous': 504, 'sharp': 497, 'slightly bumpy': 497, 'gentle': 497, 'distinct edges': 488, 'cool and dry': 480, 'uneven': 476, 'minimal give': 472, 'slightly cool': 450, 'rectangular': 449, 'synthetic and artificial feel': 444, 'rough and grainy': 444, 'bumpy texture': 442, 'slick': 430, 'slight resistance': 424, 'firm fingertip touch': 421, 'small grooves': 420, 'hard and unyielding surface': 419, 'edges': 414, 'slightly sharp edges': 409, 'synthetic and artificial': 408, 'grooved texture': 405, 'lightweight': 401, 'low-pile': 399, 'uniform texture': 398, 'ticklish sensation': 396, 'slightly warm': 396, 'natural': 392, 'slightly slippery': 391, 'fine grain': 391, 'cool and solid sensation': 387, 'hollow': 371, 'slightly flexible': 360, 'plastic': 359, 'wooden texture': 353, 'natural and organic': 352, 'uneven surface': 343, 'fibrous texture': 343, 'fabric-like': 337, 'low-friction': 334, 'transparent': 333, 'unforgiving': 330, 'thick': 322, 'light and airy': 321, 'solid resistance': 312, 'solid surface contact': 302, 'soft and smooth': 301, 'rough and coarse surface': 301, 'warm and inviting texture': 300, 'solid and sturdy feel': 296, 'furry': 288, 'strong skin brush': 286, 'slightly sticky': 284, 'coarse texture': 281, 'absorbent': 274, 'soft and fuzzy': 270, 'rough and solid': 269, 'solid and sturdy structure': 269, 'soft touch': 269, 'luxurious': 266, 'rough and dry': 266, 'mossy': 265, 'plastic-like': 261, 'uniform surface': 261, 'small indentations': 257, 'with a fine grain': 254, 'warm and dry': 253, 'slightly splintery': 252, 'distinct edges and corners': 249, 'distinct grain pattern': 249, 'natural and organic feel': 247, 'and hard': 238, 'woven': 237, 'porous': 235, 'soft and flexible': 232, 'soft and fluffy': 230, 'organic': 228, 'papery': 227, 'airy': 224, 'solid feel': 219, 'alive': 211, 'pliable': 209, 'tactile': 207, 'firm texture': 206, 'high friction': 204, 'tightly-woven': 202, 'flat surface': 202, 'unyielding surface': 201, 'heavy and dense': 201}

\begin{figure}[!ht]
    \centering
    \includegraphics[width=\linewidth]{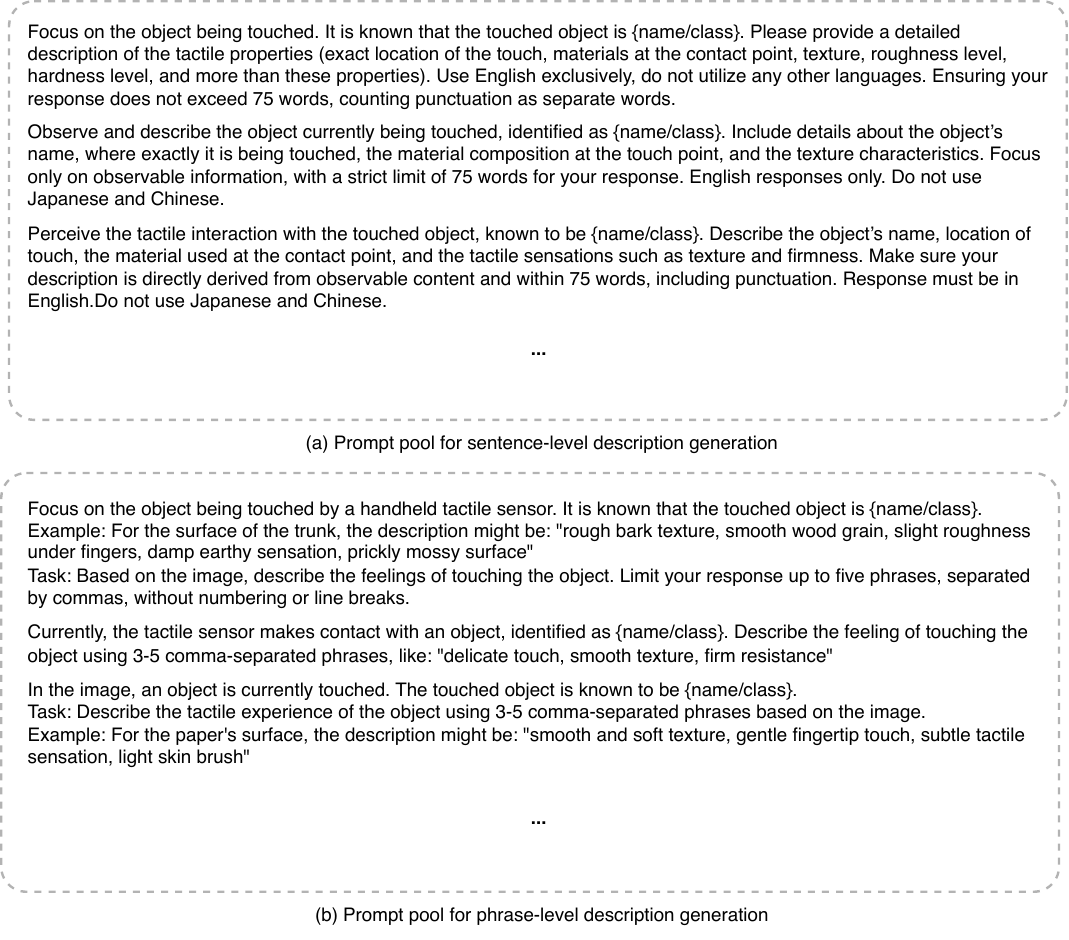}
    \caption{Prompt pools for description generation.}
    \label{fig:prompt}
    \vspace{-5mm}
\end{figure}

\section{Prompts}\label{sec:appendix_prompt}
The prompt pools we utilize are illustrated in Figure \ref{fig:prompt}, each comprising 16 prompts meticulously crafted by us, designated for either sentence-level description generation or phrase-level description generation.

\section{Model Hyperparameters}
\label{sec:appendix_hyper}
We train our model (\ie, \model) for 12 epochs using the AdamW optimizer \cite{loshchilov2016sgdr} with momentum $\beta_{1}, \beta_{2} =0.9, 0.98$. Our base learning rate is 2e-4. During training, we first warm up for 200 steps to reach the base learning rate, then apply cosine decay. Our model is trained with a batch size of 96 on 2 Nvidia A40 GPUs.

\section{Ethics Statement}
This study adheres to the ethical principles outlined in the Declaration of Helsinki, ensuring the well-being and rights of all participants.  We provide a comprehensive informed consent document detailing the study's nature, objectives, potential benefits and risks, and any foreseeable discomforts or inconveniences. Participation is entirely voluntary, and they have the right to withdraw from the study at any point without penalty or consequence. We take participant confidentiality and privacy very seriously. All data they collect is anonymized and securely stored in accordance with relevant laws and regulations. They also have the opportunity to ask questions and receive clarification before deciding to participate.

\section{Limitations and Future Work}\label{sec:limitations}
While our proposed pretraining method \model, achieves promising results for GelSight sensors, several limitations and future research directions deserve exploration.

\begin{itemize}
    \item Sensor Generalizability: The current dataset and pre-training approach are specifically designed for GelSight sensors. Their generalizability to other tactile sensor modalities, such as DIGIT or GelSlim, remains an open question. Future work will investigate techniques to enhance the transferability of tactile representations across diverse sensor types. We intend to develop techniques to extend tactile representations for broader applicability, thereby improving their efficacy across a range of vision-based tactile sensors.

    \item Multimodal Fusion Techniques: This work employs contrastive learning for modality alignment. While effective, alternative fusion techniques, such as attention mechanisms, might offer further performance gains. Future research will explore these possibilities to potentially improve the quality and robustness of the learned multimodal representation.

    \item Downstream Task Exploration: The current evaluation focuses on touch classification and grasp stability prediction. Investigating the efficacy of the learned tactile representation on a wider range of downstream tasks, such as object identification, would provide valuable insights into its versatility. Additionally, exploring applications in real-world robotic scenarios would showcase the practical value of this pretraining approach.

\end{itemize}

\section{Broader Impact}\label{sec:broader_impact}
The development of effective and generalizable tactile representation learning techniques has the potential to significantly impact various scientific and technological fields. Here, we discuss some of the potential broader impacts of our work:

\textbf{Advancements in Robotics:} Our research contributes to the Robot Grasping task. This may potentially contribute to the development of more intelligent robots capable of interacting with the physical world through touch. This can lead to robots with improved grasping capabilities, enhanced object manipulation skills, and a better understanding of material properties. Such advancements can benefit various sectors.
For example, robots with improved tactile perception can perform more delicate assembly tasks, handle fragile objects with greater care, and automate quality control processes in manufacturing facilities.

\textbf{Multimodal Understanding:} The advancement of multimodal understanding techniques, encompassing the ability to process and interpret information from various modalities like vision, language, and touch, carries far-reaching implications across several domains. Multimodal understanding can empower assistive technologies for people with disabilities. This could improve the quality of life for individuals with visual impairments.

\textbf{Human-Robot Collaboration:} By enhancing robot touch capabilities, we pave the way for more effective collaboration between humans and robots. This can enable robots to assist humans in tasks requiring delicate manipulation or complex object handling, leading to increased productivity and efficiency in various workplaces.

Overall, the development of advanced tactile representation learning holds significant promise for various scientific and technological advancements.

\end{document}